%% file: 4368.tex
\documentclass[runningheads]{llncs}
\usepackage{graphicx}
\usepackage{comment}
\usepackage{amsmath,amssymb} 
\usepackage{color}

\usepackage{cite}
\graphicspath{{./figs/}}

\usepackage{hyperref}
\usepackage[table]{xcolor}

\newcommand{\norm}[1]{\left\| {#1} \right\|}

\newcommand{\set}[1]{\left\{ {#1} \right\}}

\newcommand{\suchthat}{\ensuremath \; \left| \right. \;}

\newcommand{\pd}[2]{\frac{\partial #1}{\partial #2}}	

\newcommand{\R}{\ensuremath \mathbb{R}}

\newcommand{\eps}{\ensuremath{\varepsilon}}
 
\renewcommand{\epsilon}{\varepsilon}

\def\utilde#1{\mathord{\vtop{\ialign{##\crcr
				$\hfil\displaystyle{#1}\hfil$\crcr\noalign{\kern1.5pt\nointerlineskip}
				$\hfil\tilde{}\hfil$\crcr\noalign{\kern1.5pt}}}}}

\begin{document}
\pagestyle{headings}
\mainmatter
\def\ECCVSubNumber{4368}  

\title{LevelSet R-CNN: A Deep Variational Method for Instance Segmentation } 



\titlerunning{LevelSet R-CNN}
%
\author{
	Namdar Homayounfar* \inst{1,2} \quad  Yuwen Xiong* \inst{1,2} \quad Justin Liang* \inst{1} \\ \quad Wei-Chiu Ma \inst{1,3}  \quad Raquel Urtasun \inst{1,2}\\
	\small\texttt{\{namdar,yuwen,justin.liang,weichiu,urtasun\}@uber.com}
}

\authorrunning{Homayounfar et al.}
\institute{Uber Advanced Technologies Group \and
	University of Toronto \and
	MIT}

\maketitle

\input{abs}


\input{intro}

\input{related}

\input{background}

\input{model}

\input{experiments}

\input{conclusion}

%
%
\bibliographystyle{splncs04}
\bibliography{ref}
\end{document}

%% file: abs.tex

\begin{abstract}
Obtaining precise  instance segmentation masks is of high importance in many modern applications such as robotic manipulation and autonomous driving. Currently, many state of the art models are based on the Mask R-CNN framework which, while very powerful, outputs masks at low resolutions which could result in imprecise boundaries. On the other hand, classic variational methods for segmentation impose desirable global and local data and geometry constraints on the masks by optimizing an energy functional.  While mathematically elegant, their direct dependence on good initialization, non-robust image cues and manual setting of hyperparameters renders them unsuitable for modern applications. We propose LevelSet R-CNN, which combines the best of both worlds by obtaining powerful feature representations that are combined in an end-to-end manner with a variational segmentation  framework. We demonstrate the effectiveness of our approach on COCO and Cityscapes datasets. 

\end{abstract}

%% file: intro.tex

\section{Introduction}

Instance segmentation, the task of detecting and categorizing the pixels of unique countable objects in an image, is of paramount interest in many computer vision applications such as medical imaging \cite{Xu2016GlandIS}, photo editing \cite{Yao20183DAwareSM}, pose estimation \cite{Ma2019DeepRI}, robotic manipulation \cite{Fazelieaav3123} and autonomous driving \cite{zhang2016instance}. With the advent of deep learning \cite{alexnet2012} and its tremendous success in object classification and detection tasks \cite{Redmon2015YouOL, fastrcnn2015, renNIPS15fasterrcnn}, the computer vision community has made great strides in instance segmentation  \cite{Uhrig2016PixelLevelEA,panet,ArnabT17,bai2017deep,upsnet,Huang2019MaskSR,Chen2019TensorMaskAF}. 

Currently, the prevailing instance segmentation approaches are based on the Mask R-CNN \cite{he2017mask} framework which detects and classifies objects in the image and further processes each instance to produce a binary segmentation mask.
While achieving impressive results in many benchmarks, the predicted masks are produced at a low resolution and label predictions are independent per pixel, which could result in imprecise boundaries and irregular object discontinuities. 

In contrast, traditional variational segmentation methods \cite{kass1988snakes,chan2001active, caselles1997geodesic} are explicitly designed to delineate the boundaries of objects and handle complicated topologies. They first encode desired geometric properties into an energy functional and then evolve an initial contour according to the minimization landscape of the energy functional. One seminal work in this direction is the \emph{Chan-Vese} \cite{chan2001active} level set method, which formulates the segmentation problem as a partitioning task where the goal is to divide the image into two regions, each of which has similar intensity values. Through an energy formulation, Chan-Vese can  produce good results even from a coarse initialization. However, in the real world, the photometric values may not be consistent, for example due to illumination changes and varying textures, rendering this method impractical for modern challenging applications.

With these problems in mind, we propose {\it LevelSet R-CNN}, a novel deep structured model that combines the strengths of modern deep learning with the energy based Chan-Vese segmentation framework. 
Specifically, we build our model in a multi-task setting following the Mask R-CNN framework: four different heads are utilized based on Feature Pyramid Network (FPN) \cite{Lin2016FeaturePN} to output object localization and classification, a truncated signed distance function (TSDF) as the mask initialization, a set of instance-aware energy hyperparameters, and a deep object feature embedding, as shown in Fig.~\ref{fig:model}. These intermediate outputs are then passed into a differentiable unrolled optimization module to refine the  predicted TSDF mask of each detected object by minimizing the Chan-Vese energy functional. This results in more precise object masks at higher resolutions.

We evaluate the effectiveness of our method on the challenging Cityscapes \cite{Cordts2016Cityscapes} instance segmentation task, where we achieve state-of-the-art results. We show also improvements over the baseline on the COCO \cite{lin2014microsoft} and the higher quality LVIS \cite{gupta2019lvis}  datasets. Finally, we evaluate our model choices through extensive ablation studies.

%% file: related.tex

\section{Related Work}

\paragraph{Instance Segmentation:} Current modern instance segmentation methods can be classified as being either a top down or a bottom up approach.
In a top down approach \cite{Chen2018MaskLab, Chen_2019_CVPR, cai18cascadercnn, bshapenet, Fu2019RetinaMaskLT}, region proposals for each instance are generated and a voting process is used to determine which ones to keep. Masks are predicted from these proposals to obtain the final instance segmentation output. For example,\cite{Dai2015InstanceAwareSS} uses a cascade of networks to predict boxes, estimate masks and categorize objects in a sequential manner so that the convolutional features are shared between the tasks. In \cite{2016FullyCI}, the authors use position sensitive inside/outside score maps to jointly perform detection and object segmentation. Recently, Mask R-CNN \cite{he2017mask} augments Faster R-CNN \cite{renNIPS15fasterrcnn} to achieve very strong instance segmentation performance across benchmarks. Following this paper, the authors in \cite{Huang2019MaskSR} optimize the scores of the bounding boxes to match the mask IoU, \cite{panet} adds a bottom to top aggregation path to allow for better information flow to improve the performance and \cite{upsnet, Kirillov_2019_CVPR} extend it to panoptic segmentation. In \cite{Le2017} the authors improve an initial segmentation by fine-tuning it using a recurrent unit \cite{gru2014} that mimics level set evolution. Our approach is also top down. Here we add structure to the output space of Mask R-CNN by optimizing an explicit energy functional that incorporates geometrical constraints.

The bottom up approaches \cite{kirillov2017instancecut, newell2017associative, Brabandere2017SemanticIS, Fathi2017SemanticIS, Uhrig2016PixelLevelEA} typically perform segmentation  by grouping the feature embeddings of  individual instances without any early stage object proposals. In \cite{2015ProposalFreeNF}, the authors develop a model that predicts the category confidence, instance number and instance location and use a normalized spectral clustering algorithm \cite{2001Ng} to group the instances together. In \cite{zhang2016instance, zhang2015monocular}, a CNN outputs instance labels followed by a Markov Random Field to achieve a coherent and consistent labeling of the global image. In \cite{bai2017deep}, the authors exploit a CNN to output a deep watershed energy which can be thresholded to obtain the instance components. \cite{liu2017sgn} use a sequence of neural networks to solve a sub-grouping problem that gradually increase in complexity to group pixels of the same instance. In \cite{Kendall_2018_CVPR}, the authors propose a multi task framework that as a sub-task groups pixels by regressing a vector pointing towards the object's center. \cite{Neven_2019_CVPR} are able to achieve real time instance segmentation by introducing and using a new clustering loss that encourages pixels to point towards an optimal region around the instance center. While bottom up approaches have a much simpler design than top down methods, they usually underperform in standard metrics such as average precision and recall.  In our work, we cluster feature embeddings of an instance by differentiable optimization of an energy functional embedded within a state of the art top down approach.

\paragraph{Variational Methods:} The classic pioneering active contour models (ACM) of \cite{kass1988snakes} formulate the segmentation task as the minimization of an energy functional w.r.t. an explicit contour parametrization of the boundaries. This energy functional is comprised of a data term that moves the contours to areas of high gradient in the image. Furthermore, it regularizes the contour in terms of its smoothness and curvature. The shortcomings of ACM is that it is sensitive to initialization and requires heuristics such as re-sampling of points to handle changes of topology of the contour. The Level Set frameworks of \cite{osher1988fronts, dervieux1980finite} overcome these challenges by formulating the segmentation task as finding the zero-level crossing of a higher dimensional function. In this framework, the contours of an object are implicitly defined as the zero crossing of an embedding function such as the TSDF. This eliminates the need for heuristics to handle complicated object topologies \cite{cremers2007review}. In this work, we build upon the level set framework put forward by Chan and Vese \cite{chan2001active} where we exploit neural networks to learn robust features and optimization schedules from data.

In recent years, several works have explored combining these classical variational methods with neural networks. In the context of building segmentation from aerial images,  CNNs have been deployed to output the energy terms used to evolve an active contour and develop a deep structured model that can be learned end-to-end \cite{Marcos2018LearningDS, Cheng_2019_CVPR}. In \cite{gur2019end, ling2019fast}, the authors predict the offset to an initial circle to obtain object polygons and use a differentiable renderer to compare with the ground truth mask in the presence of a ground truth bounding box. However, they are not minimizing an explicit energy functional.  These works focus on a simpler setting than us where detection is eschewed in favor of using ground truth boxes and a dedicated neural network for segmentation. Moreover, they parameterize the output space with explicit polygons which are not able to handle multi component objects without heuristics. In our work, we tackle the full instance segmentation setting with a single backbone and also use implicit level sets that can naturally handle complicated topologies without heuristics.

In the context of implicit contours, certain works have explored leveraging level sets in neural networks either as a post processing step to obtain ground truth data, or as  a loss function for deep neural networks. In a semi-supervised setting, initial masks have been predicted for unlabeled data then further refined with level set evolution to create a quasi ground truth label \cite{tang2017deep}. The authors in \cite{hu2017deep, kim2019cnn, Chen2019LearningAC} employ level set energies as a loss function for saliency estimation and semantic segmentation.  In contrast, we employ level set optimization as a differentiable module within a deep neural network. In the experimental section, we evaluate the efficacy of using a level set loss function for the the task of instance segmentation. The closest work to ours is \cite{wang2019object}, where the authors embed a different level set optimization framework within a neural network for the task of annotator in the loop foreground segmentation. There are several key differences: (i) The energy  formulation is different,  whereas their work is built upon the edge based method of \cite{caselles1997geodesic} to push the contour to the boundaries, we exploit the region based approach of \cite{chan2001active} which imposes uniformity of object masks. (ii) their setting requires ground truth object bounding boxes to output the features used in the level set optimization, while we embed the optimization within Mask R-CNN to build on top of shared features. Note that our setting is much more challenging. In the experimental section, we extend their method to the setting of instance segmentation and compare to our proposed model.

%% file: background.tex

\section{Overview of Chan-Vese Segmentation}
\label{subsec:cv}

In this section we provide a brief overview of the classic Chan-Vese  level set segmentation method \cite{chan2001active}, which we  later combine in a differentiable manner with Mask R-CNN.  Chan-Vese is a  region based segmentation approach which is capable of segmenting objects with complex topologies, e.g.,  holes and multiple components.  
This method operates globally on image intensities and is not dependent on local well-defined edge information. At a high level, Chan-Vese partitions an image to foreground and background segments by minimizing an energy functional that encourages regions to have uniform intensity values.

Let $I$ be an image defined on the image plane $\Omega \subset \R^2$. Suppose $I$ contains only one object that we wish to segment. Let $\phi: \Omega \to \R$ be the truncated signed distance function (TSDF) to the boundaries of this object taking positive values inside the object and negative outside. Let the curve $C$ correspond to, possibly multi-component, boundaries of this object. The curve $C$ can implicitly be defined as the zero crossing of $\phi$, i.e., $C = \set{x \in \R^2 \suchthat \phi(x) = 0}$.

The core idea is to evolve an initial TSDF $\phi_0$ by minimizing an energy functional $E$ such that the zero crossing $C$ of the minimizer coincides with the object boundaries. 
In the Chan-Vese \cite{chan2001active} framework, the energy functional is defined as:
\begin{align}
E(\phi, &c_1, c_2) = \lambda_1 \int_{\Omega} \norm{I(x) - c_1}^2 H(\phi(x)) dx \nonumber
\\ &+   \lambda_2 \int_{\Omega} \norm{I(x) - c_2}^2 (1-H(\phi(x))) dx 
							+	 \mu \int_{\Omega} \delta(\phi(x))\norm{\nabla \phi(x)} dx \label{eq:E_cv}
\end{align}      
where $H$ and $\delta$ are Heaviside and Dirac delta functions respectively.
The first two terms encourage the image intensity values inside and outside of the object to be close to constants $c_1$  and $c_2$ respectively. These terms impose a partitioning of the  image to two regions of similar intensity values. The last term regularizes the length of the zero level set $C$.  The parameters $\mu, \lambda_1$ and $\lambda_2$ are  positive global hyperparameters that regulate the contribution of each energy term.

The minimization of Eq. (\ref{eq:E_cv}) is achieved by  alternatively optimizing the function $\phi$ and the constants $c _1$ and $c_2$. In particular, by holding $\phi$ fixed, the minimizer of Eq. (\ref{eq:E_cv}) w.r.t. $c_1$ and $c_2$ is given by:

\begin{align}
c_1(\phi) = \frac{\int_{\Omega} I(x) H(\phi(x)) dx}{\int_{\Omega}H(\phi(x)) dx} \hspace*{1em},\hspace*{1em}
c_2(\phi) = \frac{\int_{\Omega}I(x)( 1- H(\phi(x))) dx}{\int_{\Omega}(1-H(\phi(x))) dx} \label{eq:c1c2}
\end{align}
We thus observe that $c_1$ and $c_2$ correspond to the average of the intensity values inside and outside of the object respectively.

 Next, by holding $c_1$ and $c_2$ fixed and introducing an artificial time constant $t \ge 0$, we compute the functional derivative of $E$ w.r.t. $\phi$:
 \begin{align}
 \pd{\phi(\eps)}{t} = \delta_{\eps}(\phi)\Big ( \mu \text{div}(\frac{\nabla \phi}{\norm{\nabla \phi}})  - \lambda_1\norm{I- c_1}^2   + \lambda_2 \norm{I - c_2}^2 \Big ) \label{eq:dphi_dt}
 \end{align} 
where \emph{div} is the divergence operator, $\nabla$ is the spatial derivative and we have used a soft version of $H$ and $\delta$ defined as:
 \begin{align}
 H_\eps(z) = \frac{1}{2}\Big(1 + \frac{2}{\pi}\arctan(\frac{z}{\eps}) \Big)  \hspace*{1em},\hspace*{1em} 
 \delta_\eps(z) = \frac{1}{\pi}\cdot\frac{\eps^2}{\eps^2 + z^2} \label{eq:H_dirac_soft}
 \end{align}
Finally, the update step of $\phi$ is given by: 
\begin{align}
\phi_{n} = \phi_{n-1} + \Delta t  \pd{\phi(\eps)}{t} \label{eq:phi_gd}  
\end{align}
The alternating optimization is repeated for $N$ iterations. This procedure draws similarities to clustering techniques such as K-Means, where the optimization involves alternating  assignments and cluster center computations.

While Chan-Vese segmentation is mathematically elegant and powerful, working directly on image intensities is not robust due to factors such as lighting, different textures, motion blur or backgrounds that have similar intensities to the foreground. Moreover, the energy and optimization hyperparameters such as $\mu, \lambda_1$ and $\lambda_2$, that balance the energy terms, and $\eps$ and $\Delta t$ that regulate the gradient descent have to be manually adjusted depending on the image and domain. Furthermore, different objects have different optimal hyperparameters as their appearance and resolution might be very different. As a consequence, this method is not used in modern segmentation algorithms.  In this paper, we leverage the power of deep learning to learn high dimensional object representations where the representations of pixels of the same object instance cluster together. We also learn complex inference schedules via data dependent adaptive hyperparameters for the energy terms and the optimization.

%% file: model.tex

\section{LevelSet R-CNN}

In this section, we develop a deep structured model for the task of instance segmentation by combining the strengths of  modern deep neural networks with the classical continuous energy based Chan-Vese \cite{chan2001active} segmentation framework. In particular, we build on top of Mask R-CNN  \cite{he2017mask}, which has been widely adopted   for object localization and segmentation.  However, the masks it produces suffer from low resolution resulting in segmentations that roughly have the right shape but are not precise. Moreover, pixel predictions are independent and there is no explicit mechanism encouraging neighboring pixels to have the same label. On the other hand, the Chan-Vese segmentation framework provides an elegant mathematical approach for global region based segmentation which encourages the pixels within the object to have the same label. However, it suffers in the presence of objects with different appearances within the instance,  as  it relies on non-robust intensity cues. In this paper we take the best of both worlds by combining these two paradigms.  

We build on top of Mask R-CNN to first locate the objects in the image from the detection branch. Next, for each detected RoI corresponding to that object, we predict an initial TSDF $\phi_0$, the set of hyperparameters $\set{\mu, \lambda_1, \lambda_2}$ for the energy terms and $\set{\eps, \Delta t}$ for the optimization, and finally a deep feature embedding $F$ that will replace the image intensities in \eqref{eq:E_cv}. 
These predictions in turn will be fed into the Chan-Vese module where the costs are created and the optimization is unrolled for $N$ steps as layers of a feedforward neural network. This module will output an evolved TSDF $\phi_N$ for each object such that its zero crossing corresponds to the boundaries of this object. 

In what follows, we first describe how we build on top of Mask R-CNN, and then discuss  how inference is performed in the deep Chan-Vese module. Finally, we will describe how learning is done in an end-to-end manner. 

\begin{figure*}[t!]
	\centering
	\includegraphics[width=0.92\linewidth]{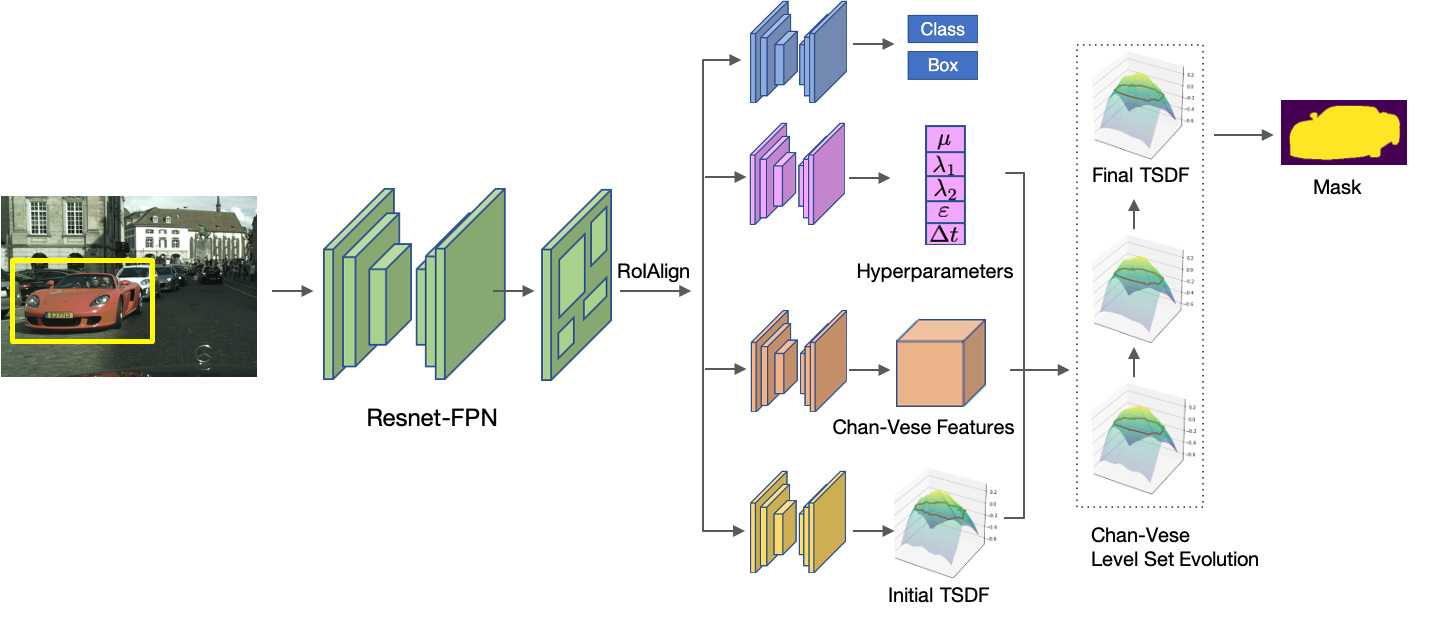}

	\caption{\textbf{LevelSet R-CNN for Instance Segmentation:} We build on top of Mask R-CNN to first detect and classify all objects in the image. Then for each detection, the corresponding RoI is fed to a series of convolutions to obtain a truncated signed distance function (TSDF) initialization, a deep feature tensor, and a set of instance aware adaptive hyperparameters. These in turn are inputted into an unrolled Chan-Vese level set optimization procedure which outputs a final TSDF. We obtain a mask by applying the Heaviside function to the TSDF.}
	\label{fig:model}
\end{figure*}

\subsection{LevelSet R-CNN Architecture} 

Here we describe the specifics of the backbone and the additional heads of our model that provide the necessary components for the Chan-Vese optimization. The model architecture is presented in Fig. \ref{fig:model}.

\paragraph{Backbone, Object Localization and Classification:} As our shared backbone we employ a Residual Network \cite{He2015DeepRL} augmented with an FPN \cite{Lin2016FeaturePN} and an RPN \cite{renNIPS15fasterrcnn} that provides object region proposals. For object localization and classification, we maintain the original head structure of  Mask R-CNN where RoIs are passed through a series of fully connected layers to output bounding box coordinates and object classification scores. 
Next, using  \emph{RoIAlign}  \cite{he2017mask} we  extract features from the backbone that are further processed by the \emph{initial TSDF head}, \emph{hyperparameter head},  and the \emph{Chan-Vese features head}. We denote the features corresponding to a RoI by $r_m$ for $m \in \set{1, \dots, M}$.  We refer the reader to the supplementary material for the exact architectural details.

\paragraph{Initial TSDF Head:}
We replace the binary output of the mask head of Mask R-CNN to produce a TSDF output instead. Specifically, each pixel of the $28 \times 28$ output provides the signed $\ell_2$ distance to the closest point on the object boundary. Furthermore, we threshold the values to a fixed symmetric range and normalize to $[-1, 1]$. This output is upsampled to $112 \times 112$ and used as the initial TSDF $\phi_{0}(r_m)$ in Eq. (\ref{eq:E_cv}).

\paragraph{Hyperparameter Head:} Each object instance could benefit from an adaptive set of hyperparameters for the energy terms and optimization steps. To achieve this, we output $\lambda_{1}(r_m)$ and  $\lambda_{2}(r_m)$ to adaptively balance the influence of the foreground and background pixels in Eq. \ref{eq:E_cv} for the object. We also predict $\mu(r_m)$ to regulate the length of its boundary. For the optimization hyperparameters, we output a separate $\eps_n(r_m)$  for each of the $N$ iterations. As  shown in Eqs. \ref{eq:dphi_dt} and \ref{eq:phi_gd}, larger values of $\eps_n(r_m)$ update the TSDF $\phi_n$ more globally and smaller values focus the evolution on the boundaries.  Similarly, we output $N$ step sizes $\Delta t_n(r_m)$ for each gradient descent step. 

To predict the above hyperparameters, we add an additional head to the RoI $r_m$ that applies a series of convolutions followed by average pooling and two fully connected layers to output a vector of dimension $2N+3$.   To ensure that these hyperparameters are positive, we found that applying a \emph{sigmoid} layer and multiplying by 2 works well.

\paragraph{Chan-Vese Features Head:}   The energy in Eq. (\ref{eq:E_cv}) encourages partitioning of the image based on the uniformity of image intensities $I$ inside and outside of the object. However, image intensity values can be non-regular due to many factors such as lighting, different textures, motion blur, etc. 
Hence, we map the image intensities to a higher dimensional feature embedding space which is learned such that pixels of the same instance are close together in embedding space.  
We achieve this by passing the RoI $r_m$ through a sequence of convolutions and upsampling layers to output a feature embedding $F(r_m)$ of dimension $C\times H \times W$. In our experiments we found $C=64$ and $H=W=112$ to be the most efficient in terms of memory for training and inference. The feature embedding $F(r_m)$ will replace the image intensities $I$ in Eq. \ref{eq:E_cv}.

\paragraph{Chan-Vese Optimization as a Recurrent Net:}

After obtaining the initial TSDF, the set of hyperparameters, and the Chan-Vese feature map, we optimize the following deep energy functional $E_m$ for each RoI $r_m$:
\begin{align}
E_m(\phi,& c_1, c_2) = \lambda_{1}(r_m) \int_{\Omega_m} \norm{F(r_m)(x) - c_1}^2 H(\phi(x)) dx \nonumber \\
&+   \lambda_{2}(r_m)  \int_{\Omega_m} \norm{F(r_m)(x) - c_2}^2 (1-H(\phi(x))) dx \nonumber \\
&+	 \mu(r_m) \int_{\Omega_m} \delta_{\eps}(\phi(x))\norm{\nabla \phi(x)} dx \label{eq:E_cv_deep}
\end{align}      
Note that the integration is over the image subset $\Omega_m \subset \Omega$ corresponding to $r_m$.
We perform alternating optimization of $\phi$ and $c_1, c_2$. We implement the $\phi$ update step:
\begin{align}
\phi_{n} = \phi_{n-1} + \Delta t_{n}(r_m)  \pd{\phi(\eps_{n}(r_m))}{t} \label{eq:phi_gd_deep}  
\end{align}
for $n=1, \dots, N$ as a set of feedforward layers with
\begin{align}
\pd{\phi(\eps_{n}(r_m))}{t} &= \delta_{\eps_{n}(r_m)}(\phi)\Big ( \mu(r_m) \text{div}(\frac{\nabla \phi}{\norm{\nabla \phi}}) \nonumber - \lambda_{1}(r_m)\norm{ F(r_m)- c_1}^2   \nonumber \\
&+ \lambda_{2}(r_m)\norm{F(r_m) - c_2}^2 \Big )
\end{align} 

In practice, we implement the gradient and the divergence term by using the Sobel operator \cite{sobel}  and the integration as a sum on the discrete image grid. At each update step, the constants $c_1$ and $c_2$ have closed-form updates as:
\begin{align}
c_1(\phi) = \frac{\int_{\Omega_m} F(r_m)(x) H(\phi(x)) dx}{\int_{\Omega_m}H(\phi(x)) dx}, c_2(\phi) = \frac{\int_{\Omega_m}F(r_m)(x)( 1- H(\phi(x))) dx}{\int_{\Omega_m}(1-H(\phi(x))) dx} \label{eq:c1c2_deep}
\end{align}
Here $c_1$ and $c_2$ are vectors where each element is the  average of the corresponding feature embedding channel inside or outside of the object in the ROI respectively.

\subsection{Learning} 
\label{subsec:learning}

We train our model jointly in an end-to-end manner,  as the Mask R-CNN backbone, the three extra heads, and the deep  Chan-Vese recurrent network are all fully differentiable. We employ the standard regression and cross-entropy losses for the bounding box and classification components of both the RPN and the detection/classification heads of the backbone. 
For training the weights of the initial TSDF head, the hyperparameter head and the Chan-Vese features head, we apply the following loss, which is a mix of $l_1$ and binary cross-entropy \emph{BCE}, to the initial and final TSDFs $\phi_0$ and $\phi_N$:
\[
\ell_{TSDF}(\phi_{\{0,N\}}, \phi_{GT}, M_{GT})  = \norm{\phi_{\{0,N\}}-\phi_{GT}}_1 + BCE(H_{\eps}(\phi_{\{0,N\}}), M_{GT}) \label{eq:ell_TSDF}
\]
Here $M_{GT}$ and $\phi_{GT}$ are the ground truth mask and TSDF targets. In order to apply \emph{BCE} on $\phi_0$ and $\phi_N$, similar to \cite{wang2019object} we map them to $[0,1]$ with the soft Heaviside function and $\eps = 0.1$. 
During backpropagation, the loss gradient from  $\phi_N$ flows through the unrolled level set optimization and then through the Chan-Vese features head, the hyperparameter head, and the initial TSDF head.

%% file: experiments.tex

\begin{table*}[t!]
	\setlength{\tabcolsep}{2pt}
	\centering
	\hfil
	\scalebox{0.78}{
		\begin{tabular}{c|c|c|cc|cccccccc}
			\multicolumn{1}{c|}{} & \multicolumn{1}{c|}{training data} &  \multicolumn{1}{c|}{AP$_ {\texttt{val}}$} & \multicolumn{1}{c}{AP$_{\texttt{test}}$} & AP$_{\texttt{test}}^{50}$ & \multicolumn{1}{c}{person}& \multicolumn{1}{c}{rider} & \multicolumn{1}{c}{car} & \multicolumn{1}{c}{truck} & \multicolumn{1}{c}{bus} & \multicolumn{1}{c}{train} & \multicolumn{1}{c}{mcycle} & bcycle \\ 
			\hline

			DWT \cite{bai2017deep}  & \texttt{fine} &$21.2$ &$19.4$ &$35.3$ &$15.5$ &$14.1$ &$31.5$ &$22.5$ &$27.0$ &$22.9$ &$13.9$ &$8.0$ \\
			Kendall et al. \cite{Kendall_2018_CVPR}  & \texttt{fine} &$-$ &$21.6$ &$39.0	$ &$19.2$ &$21.4$ &$36.6$ &$18.8$ &$26.8$ &$15.9$ &$19.4$ &$14.5$ \\
			Arnab et al. \cite{ArnabT17}  & \texttt{fine}  &$-$ &$23.4$ &$45.2$ &$21.0$ &$18.4$ &$31.7$ &$22.8$ &$31.1$ &\textbf{31.0} &$19.6$ &$11.7$\\
			SGN \cite{liu2017sgn}  & \texttt{fine+coarse} &$29.2$ &$25.0$ &$44.9$ 
			&$21.8$ &$20.1$ &$39.4$ &$24.8$ &$33.2$ &$30.8$ &$17.7$ &$12.4$ \\
			PolyRNN++ \cite{polygon-rnn++}  & \texttt{fine} &$-$ &$25.5$ &$45.5$ &$29.4$ &$21.8$ &$48.3$ &$21.2$ &$32.3$ &$23.7$ &$13.6$ &$13.6$ \\
			Mask R-CNN \cite{he2017mask} & \texttt{fine} &$31.5$ &$26.2$ &$49.9$ &$30.5$ &$23.7$ &$46.9$ &$22.8$ &$32.2$ &$18.6$ &$19.1$ &$16.0$ \\
			BShapeNet+ \cite{bshapenet} & \texttt{fine} & $-$ &$27.3$ &$50.4$
			&$29.7$ &$23.4$ &$46.7$ &$26.1$ &$33.3$ &$24.8$ &$20.3$ &$14.1$ \\
			GMIS  \cite{Liu2018AffinityDA}  & \texttt{fine+coarse} &$-$  & $27.3$ &$ 45.6$ &$31.5 $ &$25.2$ &$ 42.3 $ &$21.8$ &$ 37.2$ &$ 28.9$ &$18.8$ &$ 12.8$ \\
			Neven et al.  \cite{Neven_2019_CVPR}  & \texttt{fine} &$-$  & $27.6$ &$ 50.9$ &$34.5$ &$26.1$ &$ 52.4$ &$ 21.7$ &$ 31.2$ &$ 16.4$ &$ 20.1$ &$ 18.9$ \\
			PANet  \cite{panet}  & \texttt{fine} & $36.5$ &$31.8$ &$57.1$ &36.8 &\textbf{30.4} &\textbf{54.8} &27.0 &36.3 &25.5 &22.6 &\textbf{20.8} \\
			
			\hline
			\rowcolor[gray]{.95}
			Ours   & \texttt{fine} &\textbf{37.9} &\textbf{33.3} & \textbf{58.2} 
			& \textbf{37.0} & 29.2 & 54.6 & \textbf{30.4} & \textbf{39.4} & 30.2 & \textbf{25.5} &20.3 \\
			
			\hline
			\hline
			
			AdaptIS  \cite{Sofiiuk2019AdaptISAI} (F)  & \texttt{fine} & $36.3$ &$32.5$ &$52.5$ & 31.4 &29.1 &50.0 &31.6 &41.7& \textbf{39.4} &24.7 &12.1 \\
			
			SSAP \cite{Gao_2019_ICCV} (MS+F) & \texttt{fine} &$37.3$ & 32.7 &51.8 
			&$35.4$ &$25.5$ & {55.9} & \textbf{33.2} & \textbf{43.9} &$31.9$ &$19.5$ &$16.2$ \\
			
			Pan-DL \cite{cheng2019panopticdeeplab} (MS+F) & \texttt{fine} & 38.5 & 34.6 & 57.3 & 34.3 & 28.9 & 55.1 & 32.8 & 41.5 & 36.6 & 26.3 & 21.6  \\
			
			\hline
			
			\rowcolor[gray]{.95}
			Ours  (MS+F) & \texttt{fine} & \textbf{40.0} & \textbf{35.8} & \textbf{61.2} & \textbf{40.5} & \textbf{31.7} & \textbf{56.9} & 31.4 & 42.4 & 32.5 & \textbf{28.6} & \textbf{22.2}  \\
			
			\hline		
			\hline

			Mask R-CNN \cite{he2017mask}  & \texttt{fine+COCO} &$36.4$ &$32.0$ &$58.1$ &$34.8$ &$27.0$ &$49.1$ &$30.1$ &$40.9$ &$30.9$ &$24.1$ &$18.7$ \\
			
			BShapeNet+ \cite{bshapenet} & \texttt{fine+COCO} & $-$ &$32.9$ &$58.8$ 
			&$36.6$ &$24.8$ &$50.4$ &$33.7$ &$41.0$ &$33.7$ &$25.4$ &$17.8$ \\
			UPSNet \cite{upsnet} & \texttt{fine+COCO} &$37.8$ &$33.0$ &$59.7$ &$35.9$ &$27.4$ &$51.9$ &$31.8$ &$43.1$ &$31.4$ &$23.8$ &$19.1$ \\
			PANet \cite{panet} & \texttt{fine+COCO} &$41.4$ &$36.4$ &$63.1$ 
			&$41.5$ &$33.6$ &$58.2$ &$31.8$ &$45.3$ &$28.7$ &$28.2$ & 24.1 \\
			
			Pan-DL \cite{cheng2019panopticdeeplab}  & \texttt{fine+MV} & 42.5 & 39.0 & 64.0 & 36.0 & 30.2 & 56.7 & \textbf{41.5} & \textbf{50.8} & \textbf{42.5} & 30.4 & 23.7  \\

			Polytransform \cite{liang2019polytransform} & \texttt{fine+COCO} &$\bf{44.6}$ & $\bf{40.1}$ & $\bf{65.9}$ & 42.4 & $\bf{34.8}$ & 58.5 & 39.8 & 50.0 & 41.3 & 30.9 & $23.4$\\

			\hline
			
			\rowcolor[gray]{.95}
			Ours (COCO)   & \texttt{fine+COCO} & 43.3 & 40.0 & 65.7 & \textbf{43.4} & 33.9 & \textbf{59.0} & 37.6 & 49.4 & 39.4 & \textbf{32.5} & \textbf{24.9} \\
			\hline

		\end{tabular}
	}
\vspace{1mm}
	\caption{\textbf{Instance segmentation on Cityscapes val and test sets:} This table shows our instance segmentation results on Cityscape on val and test. We report models trained on Cityscapes with and without COCO/Mapillary pre-training as well the methods that use horizontal flipping (F) or multiscale (MS) inference at test time.}
	\label{tab:ap_val_test}
\end{table*}

\section{Experimental Evaluation}
In this section, we describe the datasets, implementation details and the metrics and compare our approach with the state-of-the-art. Next, we study the various aspects of our proposed approach through ablations.

\paragraph{Datasets:} We evaluate our model on Cityscapes~\cite{Cordts2016Cityscapes} and COCO \cite{lin2014microsoft} datasets. Cityscapes contains very precise annotations for 8 categories split into 2975 train, 500 validation and 1525 test images of resolution $1024\times 2048$.  The COCO dataset has 80 categories with 118k images in the \texttt{train2017} set for training and 5k images in the \texttt{val2017} set for evaluation. However, as demonstrated quantitatively by \cite{gupta2019lvis}, COCO does not consistently provide accurate object annotations rendering mask quality evaluation of a method not indicative. As such, we follow the approach of \cite{kirillov2019pointrend} and also evaluate our model on the COCO sub-categories of the validation set of the LVIS dataset \cite{gupta2019lvis} with our model trained only on COCO. Note that LVIS re-annotates all the COCO validation images with high quality masks which makes it suitable for evaluating mask improvements. We follow this protocol since the LVIS dataset has more than 1000 categories and is designed for large vocabulary instance segmentation which is still in its infancy and an exciting topic for future research.

\paragraph{Implementation Details:}  For Cityscapes, we follow ~\cite{he2017mask} and adopt multi-scale training where we resize the input image in a way that the length of the shorter edge is randomly sampled from [800, 1024]. We train the model on 8 GPUs for 24K iterations with a learning rate of 0.01, decayed to 0.001 at 18K iterations. We set the loss weights for the initial and final TSDF output to 1 and 5 in the multitask objective. 
For COCO, following ~\cite{he2017mask}, we train the model without multi-scaling on 16 GPUs for 90K iterations with a learning rate of 0.02 decayed by a factor of 10 at 60K and 80K iterations. We set the loss weights for the initial and final TSDF output to 0.2 and 1 in the multitask objective.
For both datasets, we set the weight decay as 0.0001, with mini-batch size of 8. We employ \texttt{WideResNet-38}~\cite{rotabulo2017place} on Cityscapes test set and \texttt{ResNet-50}~\cite{He2015DeepRL} in all the other experiments. For the level set optimization, we unroll the optimization for 3 steps. Finally, we simply apply the Heaviside function to the TSDF output to obtain a mask. Note that if we apply marching squares \cite{marchingcubes} to the final TSDF instead, we could obtain sub-pixel accuracy for the boundaries. However for simplicity and since the AP metric of COCO and Cityscapes requires binary masks for evaluation, we simply threshold our TSDFs using the heaviside function.

\begin{table}[t]
	\centering
	\scalebox{0.9}{
	\begin{tabular}{l|c|c|c}
		& Cityscapes AP & COCO AP & LVIS AP*   \\ \hline
		\multicolumn{1}{l|}{Mask R-CNN} 		& 32.3 &  33.8 & 35.6  \\ \hline
		\multicolumn{1}{l|}{Ours (Initial Mask)} & 35.4 & 33.7 & 35.8  \\ \hline
		\multicolumn{1}{l|}{Ours} 					&	\textbf{36.2} & \textbf{34.3} & \textbf{36.4}  \\ 
	\end{tabular}
}
	\caption{\textbf{LevelSet R-CNN vs. Mask R-CNN:} We report mask AP for both COCO and Cityscapes on the val set with \texttt{Resnet-50} backbone. We also report the federated AP, i.e. AP*, of the LVIS dataset with COCO subcategories trained only COCO. For our model we report both the initial and final mask results after optimization.} 
	\label{tab:city_coco_lvis}
	
\end{table}

\begin{table}[t]
	\centering	
	\scalebox{0.9}{
		\begin{tabular}{c|c|c|c|c}
			
			& LS Loss \cite{hu2017deep} & DELSE \cite{wang2019object} & DELSE \cite{wang2019object} + HP head & Ours \\ \hline
			Initial Mask AP & - & 34.6 & 34.9 &  \textbf{35.4}\\ \hline
			Final Mask AP & 34.3 & 34.6  & 35 & \textbf{36.2} 
		\end{tabular}
	}	
	\caption{\textbf{LevelSet R-CNN vs.  deep level set methods on the val set of Cityscapes:} Using level sets as a loss function (LS Loss) or using the geodesic level sets (DELSE). }
	\label{tab:diff_level_sets}
	
\end{table}

\begin{table}[t]
	\centering
		\scalebox{0.9}{
	\begin{tabular}{c|c|c}
		& Detach $\phi_0$ & Backprop Thru $\phi_0$ \\ \hline
		\multicolumn{1}{c|}{Initial Mask AP} & 33.0  & \textbf{35.4} \\ \hline
		\multicolumn{1}{c|}{Final Mask AP} & 34.2 & \textbf{36.2} \\ 
	\end{tabular}
}
	
	\caption{\textbf{Backproping through the initial mask} from the final TSDF loss $\ell_{TSDF}$ on the val set of Cityscapes}
	\label{tab:detach}
\end{table}

\begin{table}[t]
	\centering
		\scalebox{0.9}{
	\begin{tabular}{c|c|c|c}
		& Backbone & AF$_1$ & AF$_2$ \\ \hline
		\multicolumn{1}{c|}{Mask R-CNN \cite{he2017mask}} & \texttt{Resnet-50} & 40.2 & 57.6 \\ 
		\multicolumn{1}{c|}{Ours} & \texttt{Resnet-50} & \textbf{45.8} & \textbf{63.1} \\ \hline
		\hline
		\multicolumn{1}{c|}{Mask R-CNN \cite{he2017mask}} & \texttt{WideResnet-38} & 42.7 & 59.9 \\ 
		\multicolumn{1}{c|}{Ours} & \texttt{WideResnet-38} & \textbf{46.8}  & \textbf{64.6} \\
	\end{tabular}
}
	\caption{\textbf{Boundary metric:} on val set of Cityscapes: We evaluate the AF at thresholds of 1 and 2 pixels against Mask R-CNN across two backbones.}
	\label{tab:boundary}
\end{table}

\paragraph{Evaluation Metrics:} We report the standard AP metric of \cite{lin2014microsoft} on both Cityscapes and COCO. For LVIS, we report the federated average precision metric denoted by AP* \cite{gupta2019lvis} on the COCO subcategories.

\paragraph{Cityscapes Test:} We compare LevelSet R-CNN against published state-of-the-art (SOTA) methods on  Cityscapes in Table~\ref{tab:ap_val_test}.  LevelSet R-CNN outperforms all previous methods that are trained on Cityscapes data without test time augmentations achieving a new state-of-the-art performance by 1.5 AP over PANet \cite{panet}.  We also compare against  models that adopt multiscale (MS) and horizontal flipping (F) at test time.  We improve upon the state-of-the-art, Panoptic-Deeplab \cite{cheng2019panopticdeeplab},  by 1.2 AP. 
Next, we evaluate against models that pretrain on external datasets such as COCO \cite{lin2014microsoft} or Mapillary Vistas \cite{MVD2017}. For a fair comparison, we follow the exact setting of Polytransform \cite{liang2019polytransform} which was the state-of-the-art at the time of submission. In particular, we use a \texttt{WideResNet-38} backbone with deformable convolutions \cite{dai2017deformable} and PANet modifications \cite{panet}. We train on COCO for 270000 iterations with a learning rate of 0.02 decayed by a factor of 10 at 210000 and 250000 iterations on 16 GPUs. On Cityscapes, we finetuned for 6000 iterations on 8 GPUs with a learning rate of 0.01 decayed to 0.001 at 4000 iterations. As shown in Table \ref{tab:ap_val_test}, our performance is comparable with Polytransform.

\paragraph{AP Improvements Across Datasets:}  

In Table \ref{tab:city_coco_lvis}, we compare Mask R-CNN with our initial and final mask outputs on the validation sets of  Cityscapes, COCO and LVIS. All models employ the \texttt{Resnet-50} backbone. LevelSet R-CNN outperforms Mask R-CNN on all datasets. Note that while Levelset R-CNN was only trained on COCO and not with the precise boundaries of LVIS, it improves upon the baseline by 0.8 AP*. We also see an improvement of about 4 AP on  Cityscapes.

\paragraph{Different Deep Level Set Formulations:} To further justify our deep region based level set formulation, we compare with two different variations of  level sets applied to the task of instance segmentation. 
 \cite{hu2017deep}  use the Chan-Vese energy as a loss function for salient object detection. Here, we employ their loss for instance segmentation. In particular, we shift the mask output of Mask R-CNN by -0.5, apply the soft Heaviside and pass to the Chan-Vese energy loss function. In Table \ref{tab:diff_level_sets} we observe that LevelSet R-CNN improves the level set loss by about 2 AP. Next, we combine the deep edge based level set of \cite{wang2019object}, referred to as DELSE, with Mask R-CNN by changing the mask head to a TSDF head and adding two more heads: the velocity head  that predicts the direction to the object boundaries and the modulation head which regulates the effect of the curvature term on the object boundaries. We evaluate in two settings: 1) Similar to their work, we use hand-tuned hyperpameters and  use their exact loss functions, i.e., $L_2$ for the initial TSDF, class balanced cross entropy for the final TSDF and $L_2$ on angular domain.  2) To remove the effect of loss functions and hyperparameters  choices and provide the most fair comparison, we add our hyperparameter head to their model and use our  loss functions with the exception of having an extra loss function for the velocity head. In Table \ref{tab:diff_level_sets}, we observe that LevelSet R-CNN has a 1.2 AP improvement over the DELSE formulation. Moreover, we obtain 0.8 AP improvement over the initial mask whereas DELSE gain is 0.1 AP.

\paragraph{Passing Gradients through the Initial TSDF:} 

Table \ref{tab:detach} shows that by passing the gradient from $\phi_N$ through the initial TSDF head, we improve the AP of both the  initial TSDF $\phi_0$ and the final TSDF $\phi_N$. As an alternative we could have detached $\phi_0$ from the computation graph so that it does not take supervision from the final TSDF $\phi_N$; Passing the gradient improves $\phi_0$ by 2.4 AP and $\phi_N$ by 2 AP. Finally, we see that the AP of the initial TSDF when passing gradients is higher than the AP of the final TSDF when not passing the gradient by 1.2 AP. This suggests that the hyperparameter head, the Chan-Vese features head and the unrolled optimization, can also be used  during training for improving the performance of the mask head and discarded during inference.

\begin{figure*} [t!]
	\centering
	\setlength\tabcolsep{2pt}
	\begin{tabular}{cccc}

		\includegraphics[width=.33\textwidth]{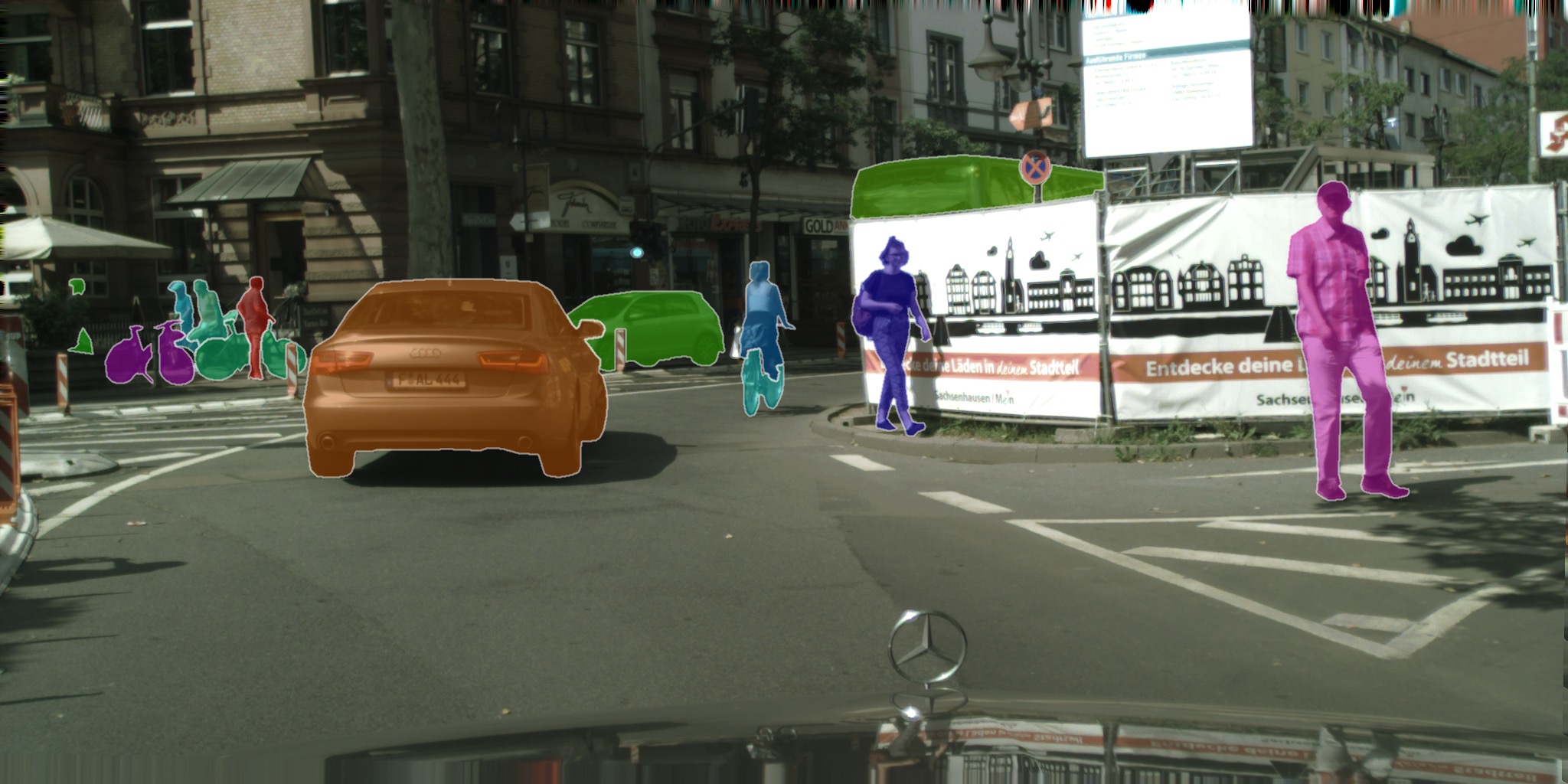} &
		\includegraphics[width=.33\textwidth]{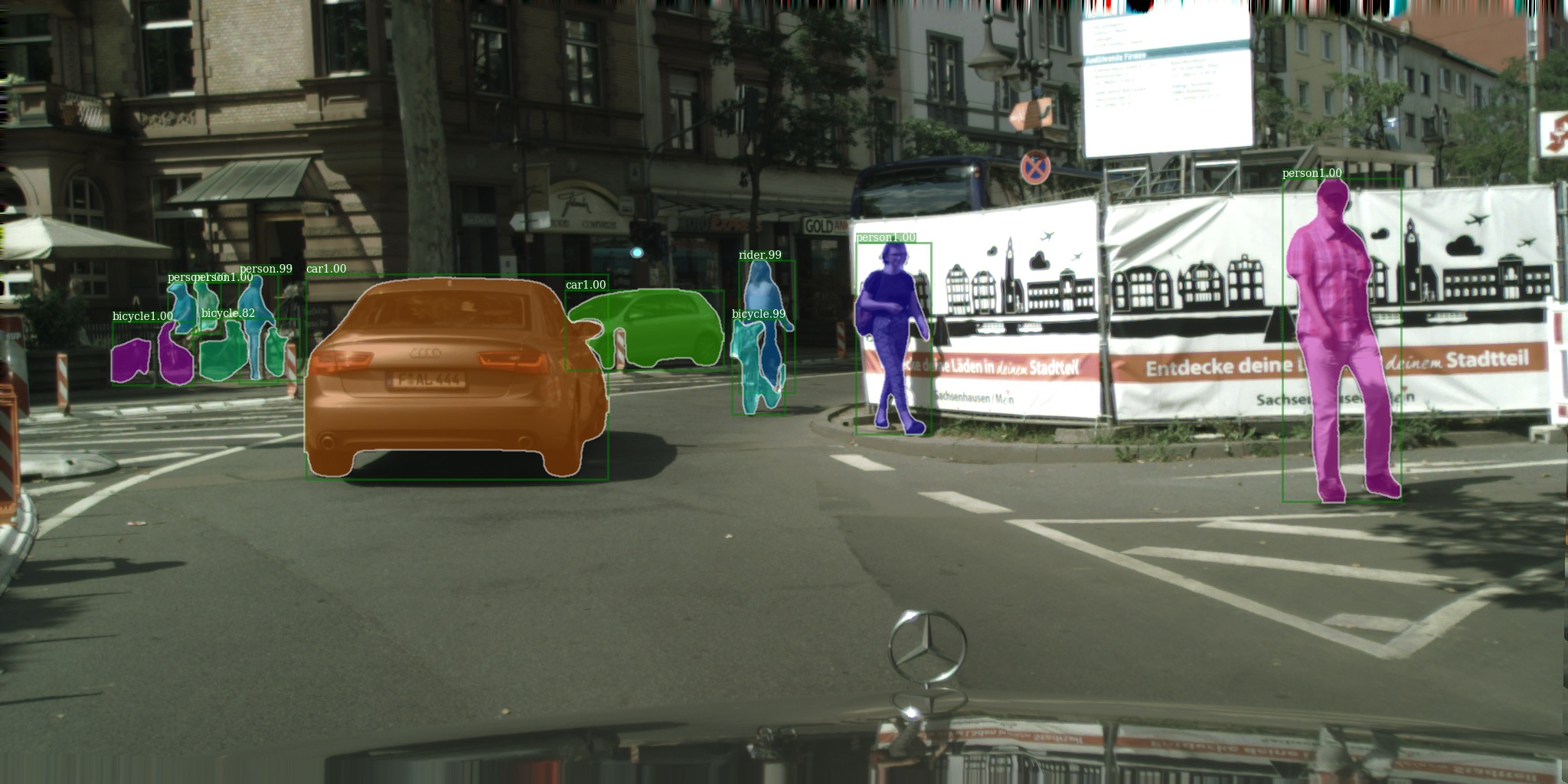} &
		\includegraphics[width=.33\textwidth]{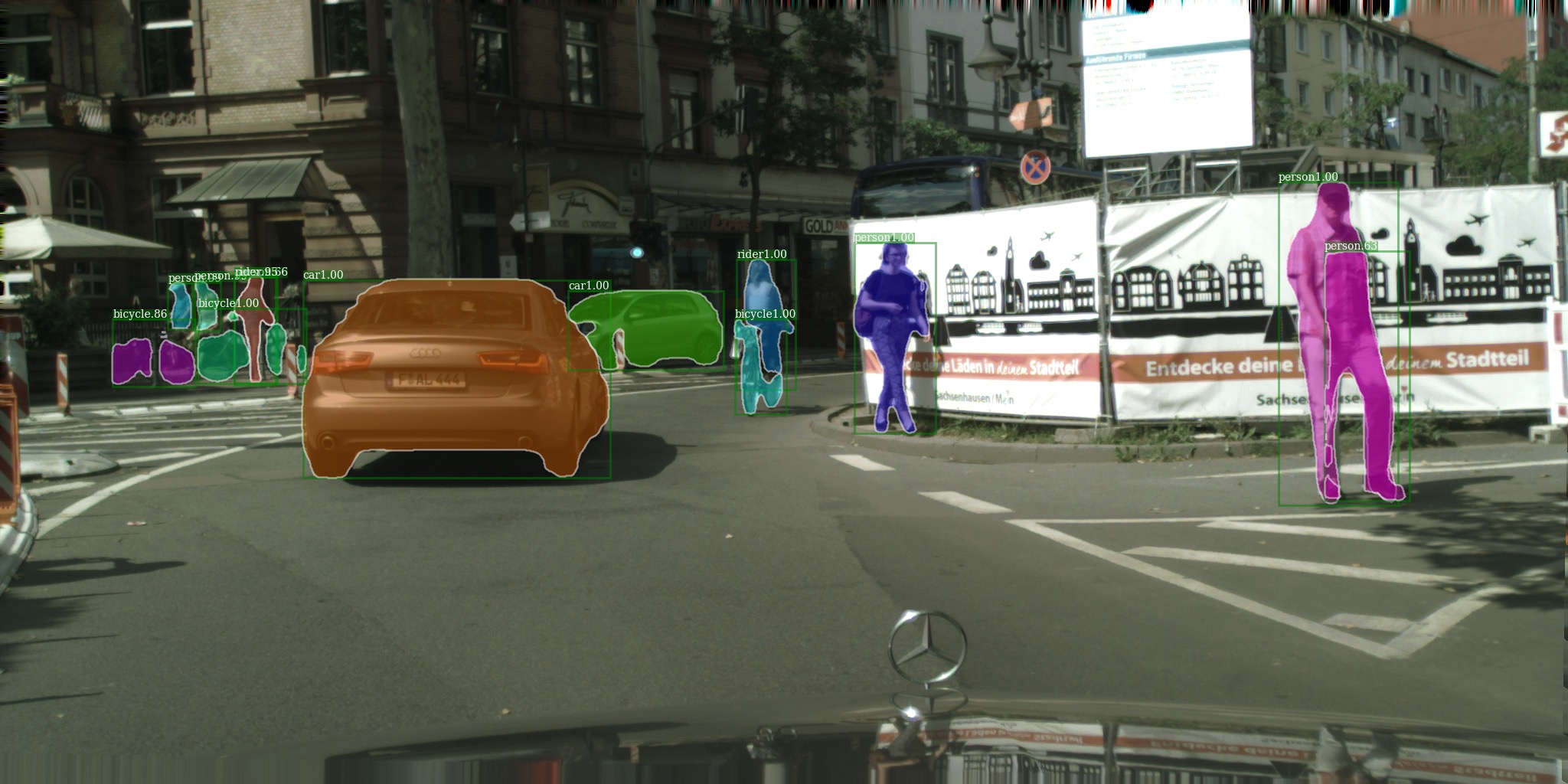}\\

		\includegraphics[width=.33\textwidth]{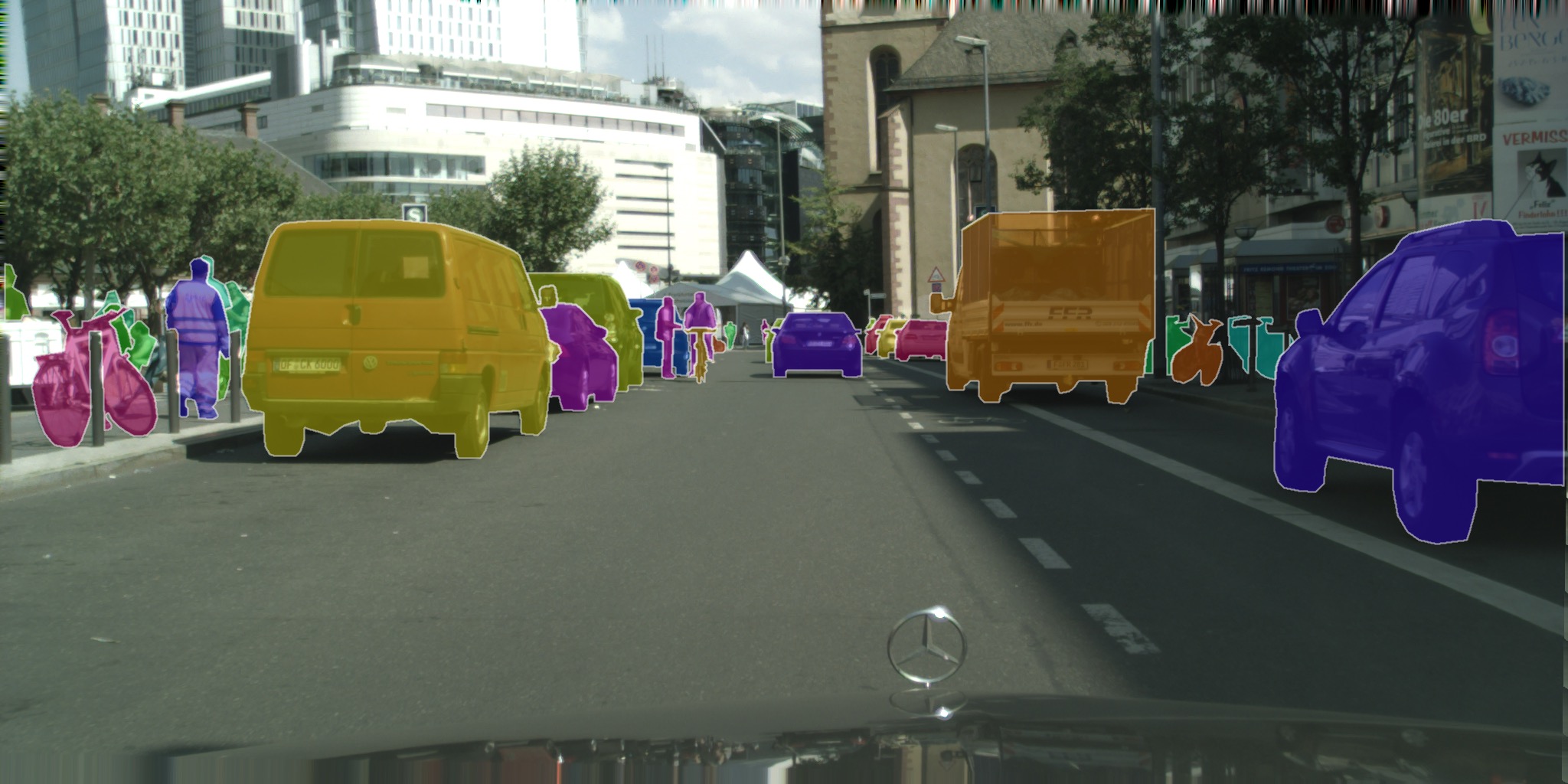} &
		\includegraphics[width=.33\textwidth]{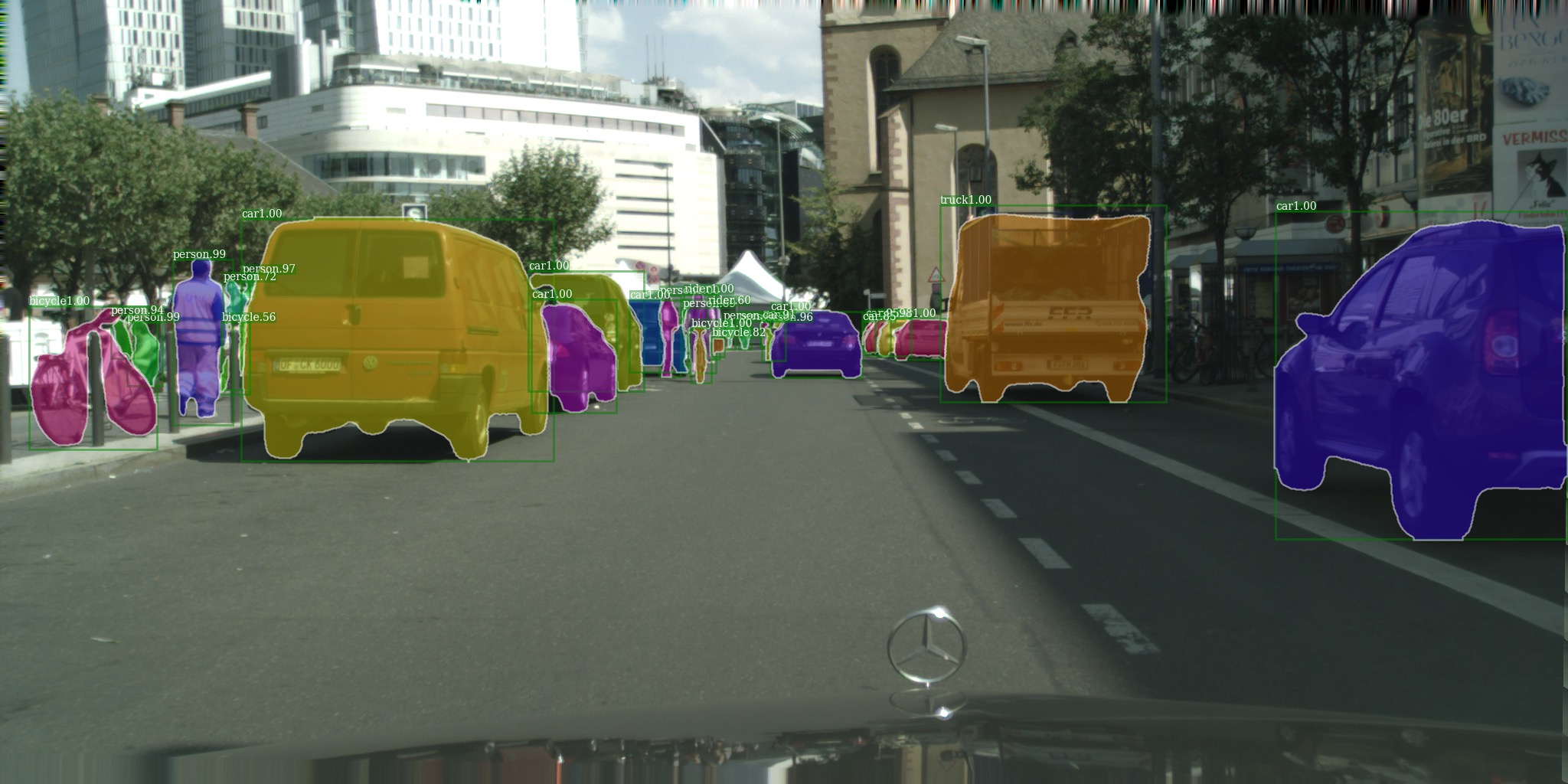} &
		\includegraphics[width=.33\textwidth]{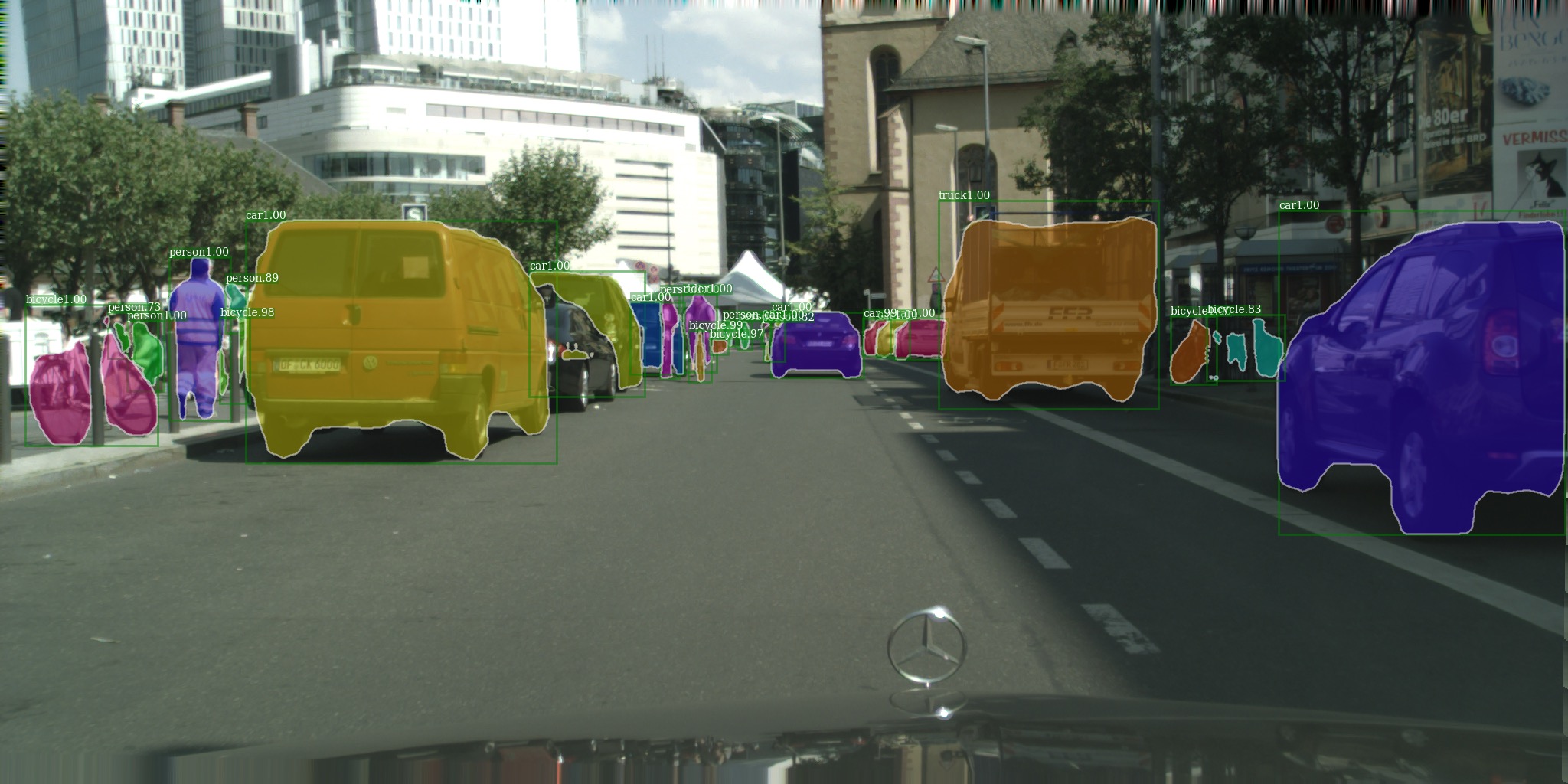}\\
		
		\includegraphics[width=.33\textwidth]{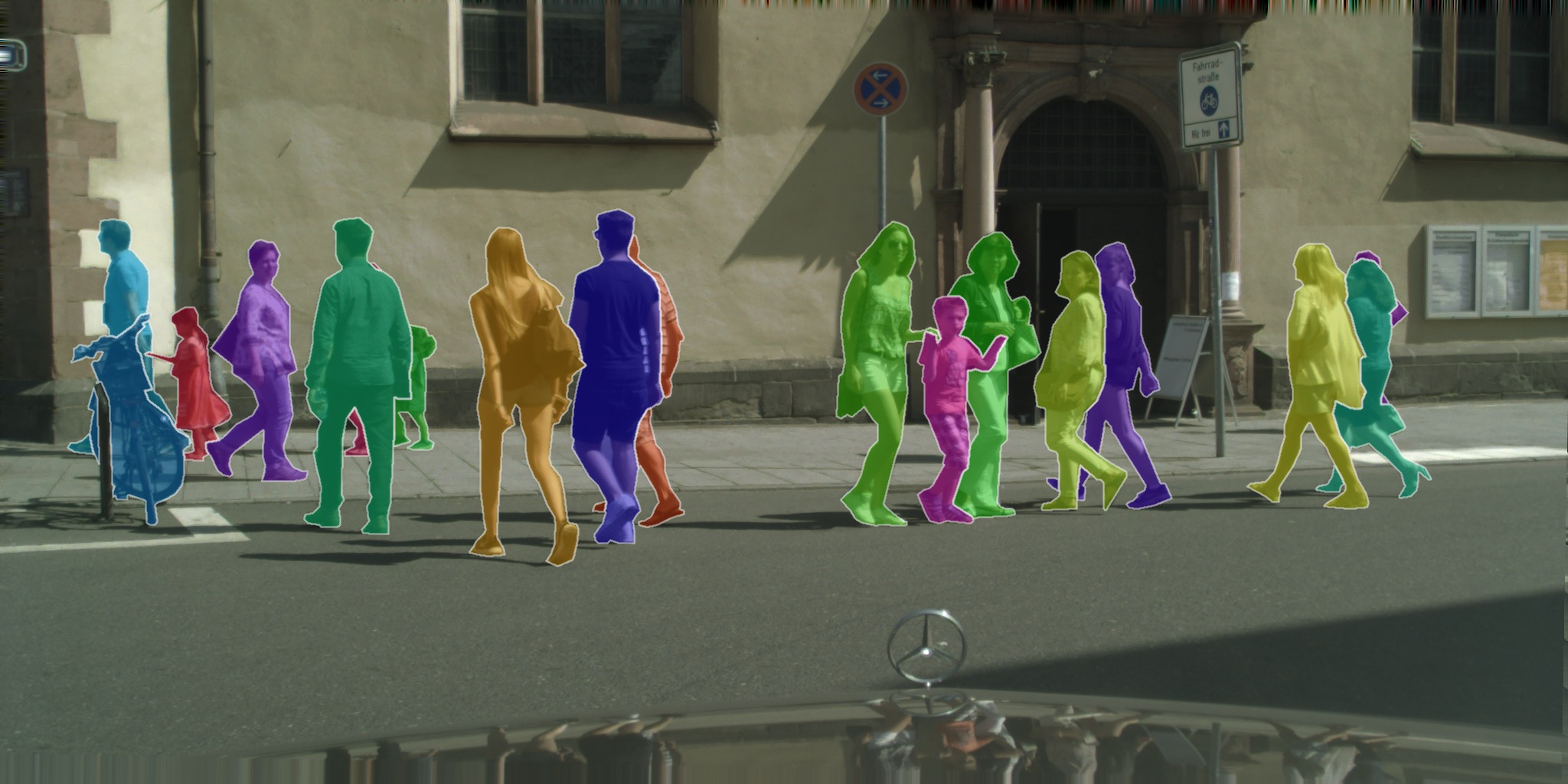} &
		\includegraphics[width=.33\textwidth]{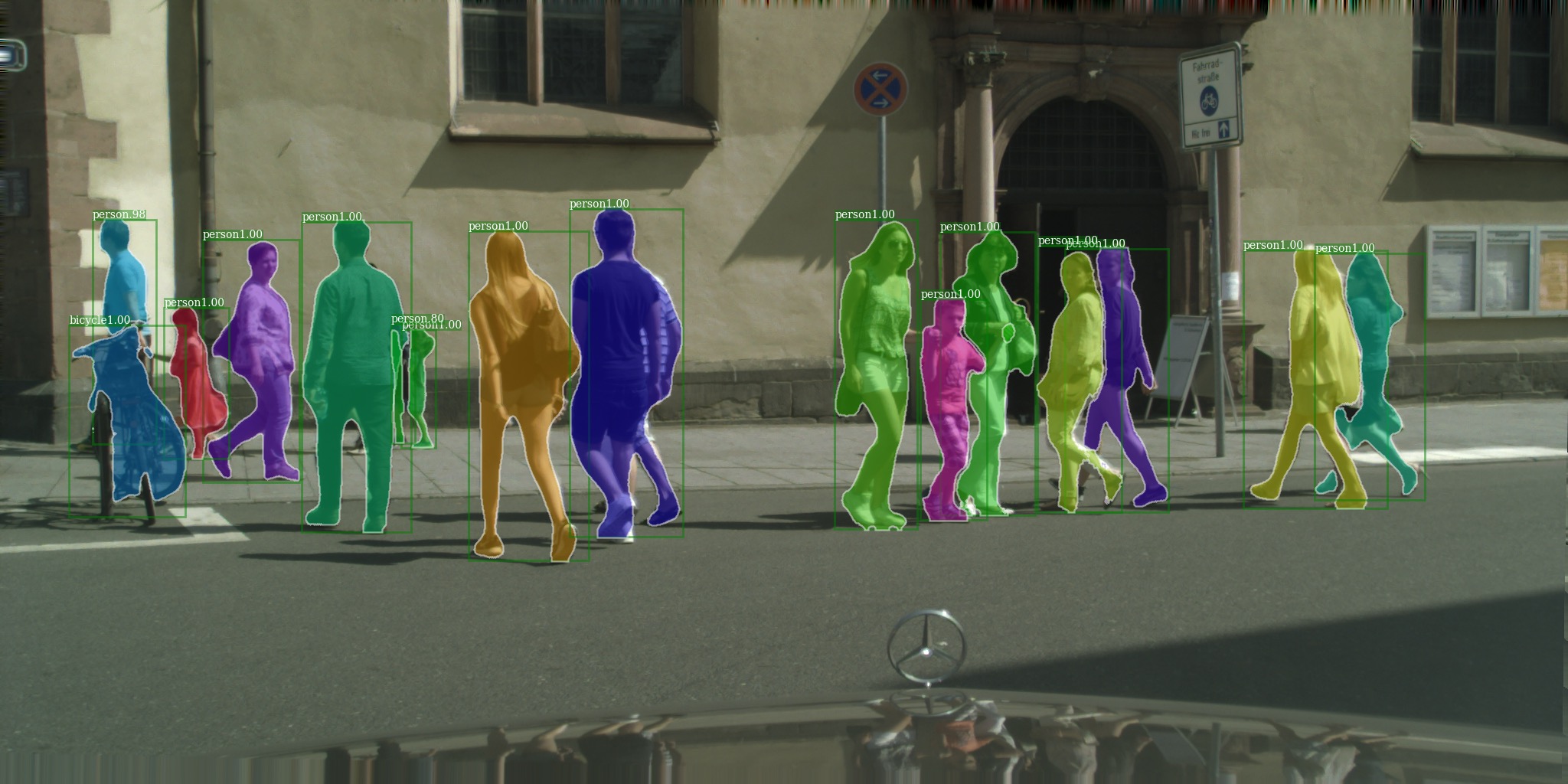} &
		\includegraphics[width=.33\textwidth]{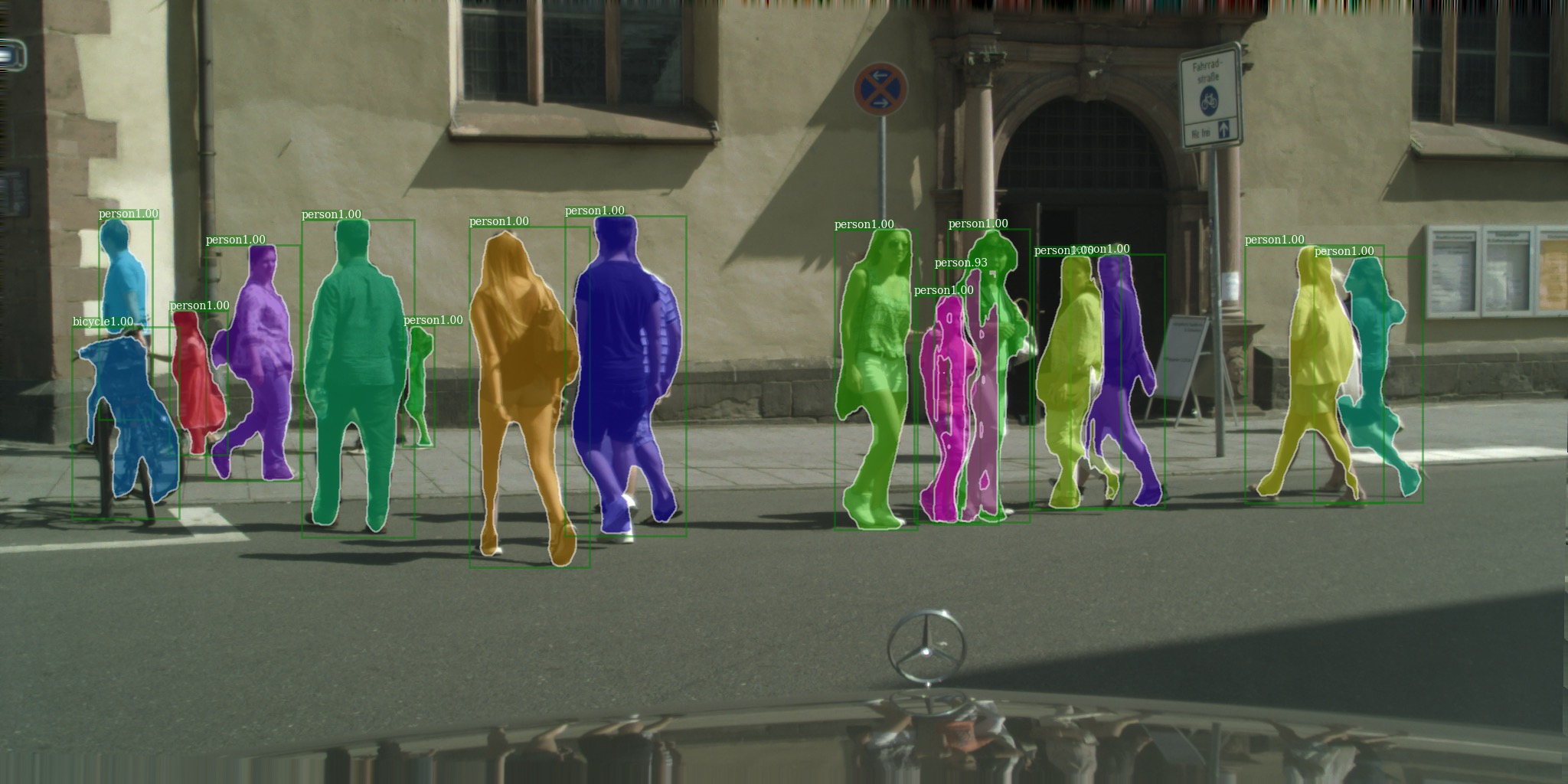}\\

		GT Instance Segmentation & Ours & Baseline
		
	\end{tabular}
	
	\caption{We showcase qualitative instance segmentation results of our model on the Cityscapes validation set. We can see that our method produces masks with higher quality when bounding box results are similar.}
	\label{fig:results_city}
\end{figure*}

\paragraph{Boundary Metric:} In addition, to evaluate the capacity of our model in improving the boundaries of objects, we adapt the boundary metric of DAVIS \cite{Perazzi2016} to our task. In particular, for a True Positive detection, we compute F1 between the prediction and ground truth boundary pixels at thresholds of 1 and 2 pixels. Similar to AP, we obtain the True Positives at IoUs in range [0.5, 0.95] at 0.05 increments. The F1s are averaged over all the classes and thresholds and are denoted by (AF$_1$) and (AF$_2$) for thresholds of 1 and 2 pixels away. In Table \ref{tab:boundary}, we observe that our method is able to improve the boundaries of the objects by at least 4 AF at each threshold compared to the baseline across the two backbones \texttt{Resnet-50} and \texttt{WideResnet-38}.

\paragraph{Mask R-CNN with Different Training Targets:} We modify the mask head of  Mask R-CNN to output a TSDF instead of a binary mask and we train with $\ell_{TSDF}$ rather than BCE as loss function. 
To understand the dependence of the Mask R-CNN performance on this TSDF target $\ell_{TSDF}$, we trained a model with only the mask head modification and without the other Chan-Vese components (i.e., the adaptive hyperparameter head and the deep Chan-Vese module and unrolled optimization). 
We obtain the same AP of 32.3 for model with $\ell_{TSDF}$ as the original Mask R-CNN. This indicates that the model improvements do not simply come from changing the loss of the mask head.

\begin{figure*} [t!]
	\centering
	\setlength\tabcolsep{0.5pt}
	\begin{tabular}{cccccc}
		
		\raisebox{15px}{\rotatebox{90}{\tiny GT}}
		\includegraphics[height=.11\textwidth]{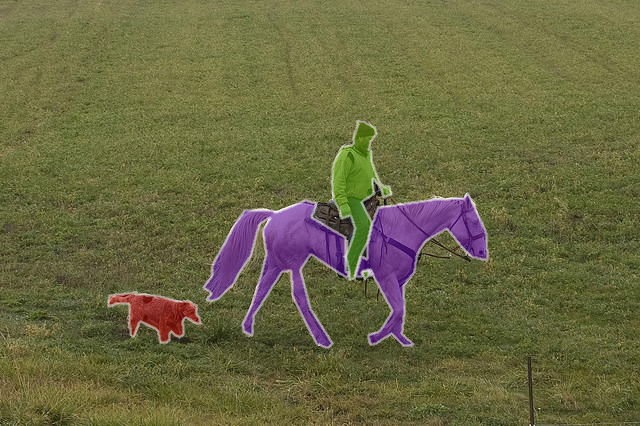} &
		\includegraphics[height=.11\textwidth]{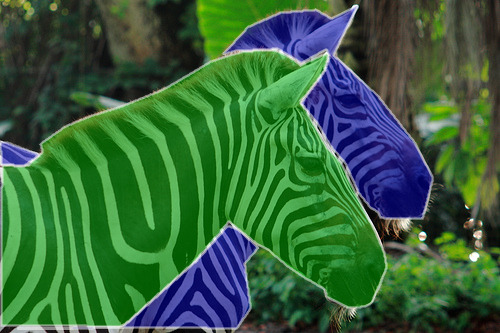} &
		\includegraphics[height=.11\textwidth]{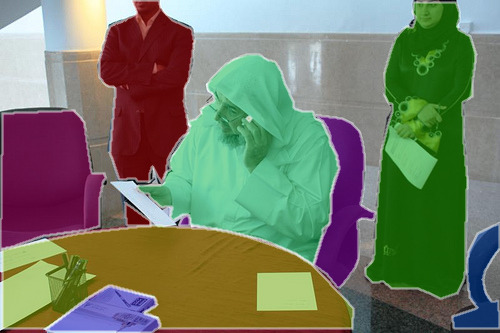} &
		\includegraphics[height=.11\textwidth]{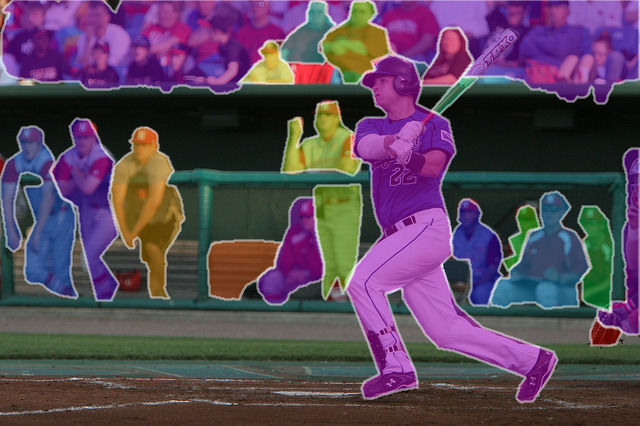} &
		\includegraphics[height=.11\textwidth]{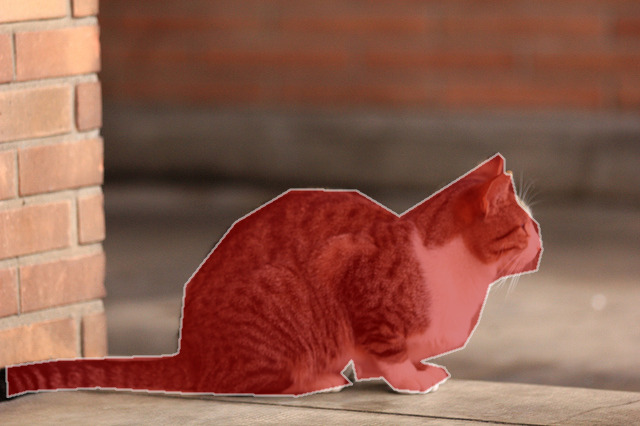} &
		\includegraphics[height=.11\textwidth]{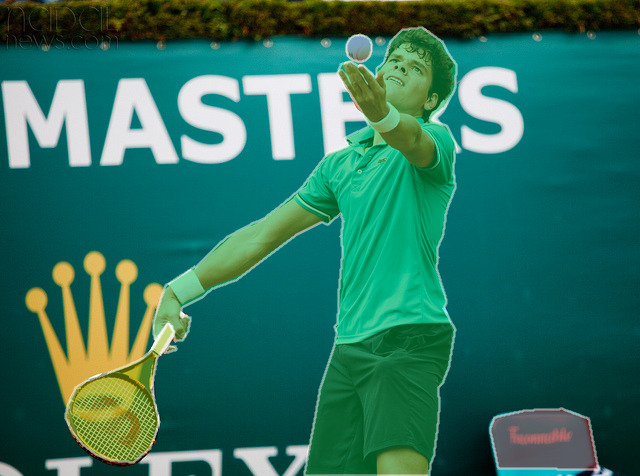} \\

		\raisebox{13px}{\rotatebox{90}{\tiny Ours}}
		\includegraphics[height=.11\textwidth]{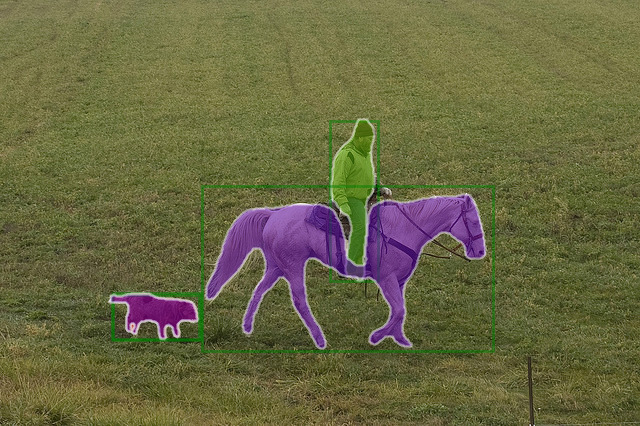} &
		\includegraphics[height=.11\textwidth]{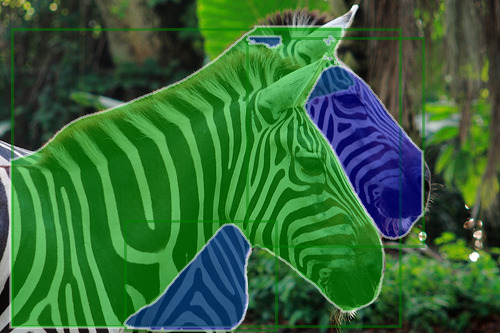} &
		\includegraphics[height=.11\textwidth]{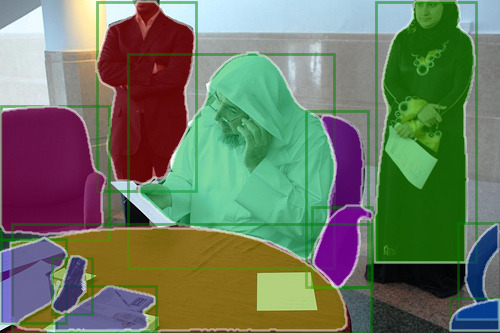} &
		\includegraphics[height=.11\textwidth]{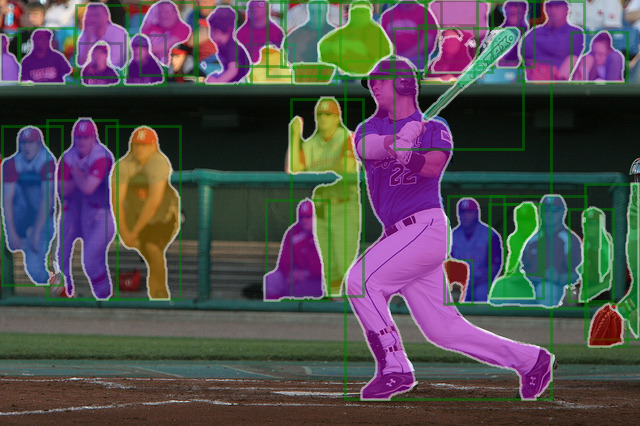} &
		\includegraphics[height=.11\textwidth]{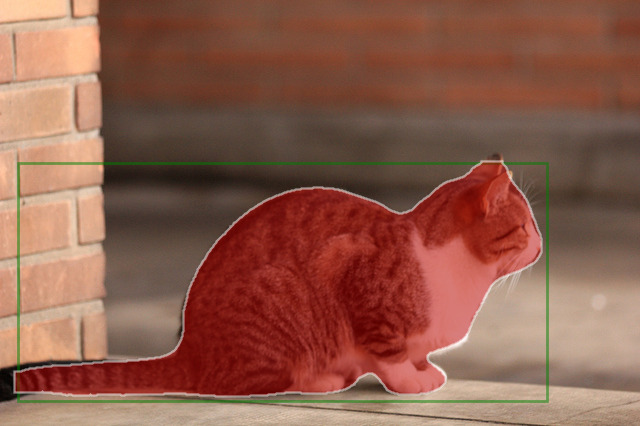} &
		\includegraphics[height=.11\textwidth]{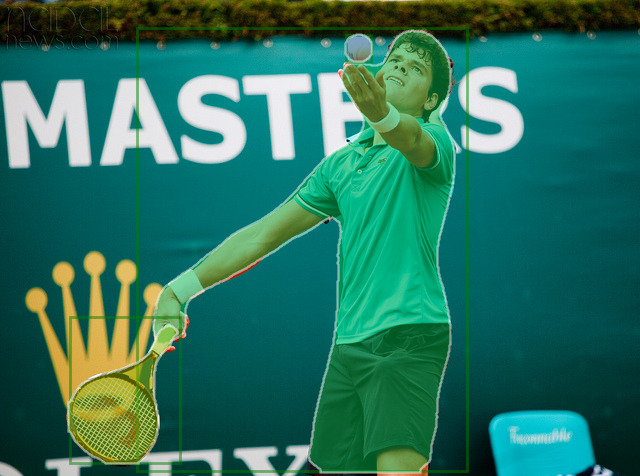} \\

		\raisebox{-1px}{\rotatebox{90}{\tiny Mask R-CNN}}
		\includegraphics[height=.11\textwidth]{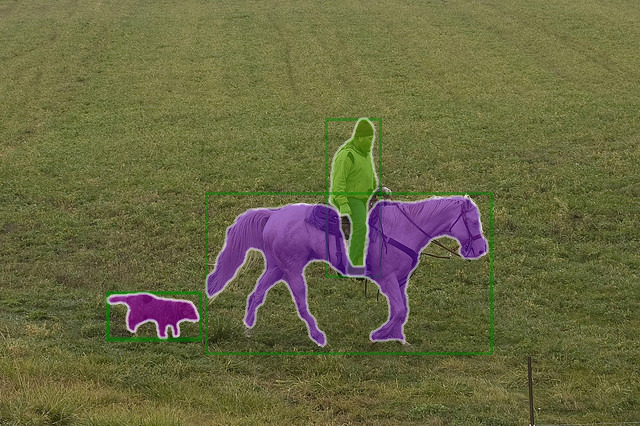} &
		\includegraphics[height=.11\textwidth]{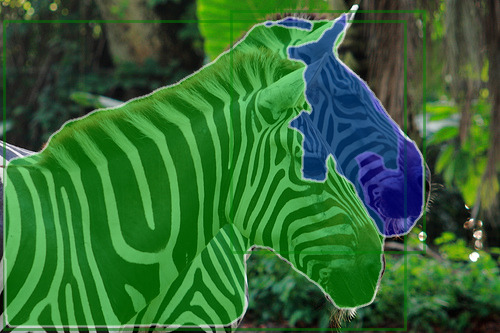} &
		\includegraphics[height=.11\textwidth]{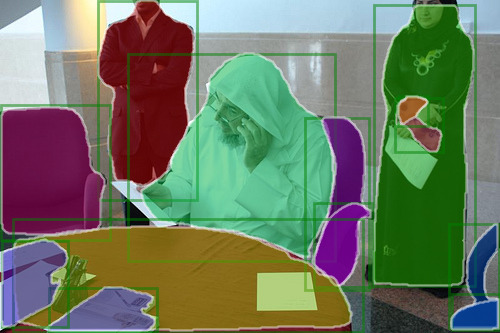} &
		\includegraphics[height=.11\textwidth]{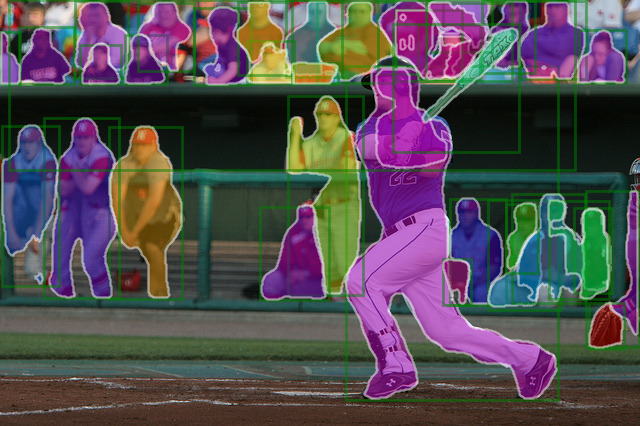} &
		\includegraphics[height=.11\textwidth]{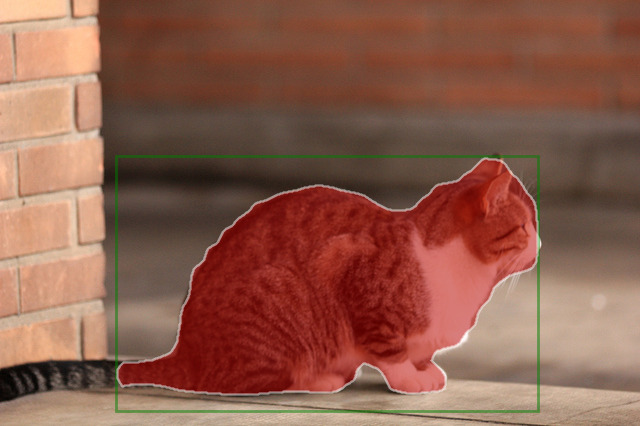} &
		\includegraphics[height=.11\textwidth]{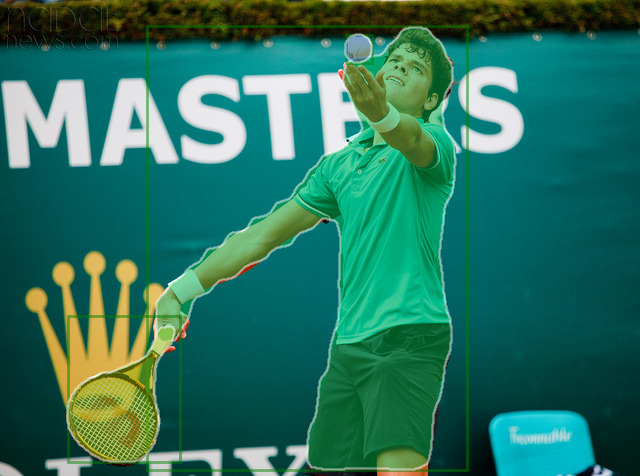} \\
		
	\end{tabular}
	
	\caption{We showcase qualitative instance segmentation results of our model on the COCO validation set.}
	\label{fig:results-coco}
\end{figure*}

\paragraph{Effect of the Hyperparameter Head:}
To verify the importance of learning adaptive hyperparameters per object instance, we perform an ablation where we remove the hyperparameter head and just learn a set of global hyperparameters for the whole dataset. The adaptive hyper parameter head achieves 36.2 AP vs. the 35.4 AP of a global set giving a boost of 0.8 AP.

\paragraph{Higher Resolution Mask R-CNN:} We evaluate whether we could improve the performance of Mask R-CNN by just increasing the resolution of the mask head. We train Mask R-CNN on Cityscapes with \texttt{ResNet-50} backbone at 112x112 resolution which is the same resolution of the Chan-Vese features and our final TSDF $\phi_N$. Interestingly the performance drops by 2.2 AP from 32.3. We hypothesize that by increasing the resolution, the ratio of non-boundary pixels vs. boundary pixels will become higher and they dominate the loss function gradients leading to worse masks. In our proposed method however, there is a global competition between the foreground/background regions to minimize the energy and hence we are able to increase the resolution.

\paragraph{Inference Time:} LevelSet R-CNN with a \texttt{ResNet-50} runs on average at 182 ms vs. Mask R-CNN at 145 ms on GTX 1080 ti on images of dimension $1024\times 2048$.

\paragraph{Qualitative Results:} As shown in Figs. \ref{fig:results_city} and \ref{fig:results-coco} we observe mask boundary and region improvements compared to the baseline.

%% file: conclusion.tex

\section{Conclusion}

In this paper, we proposed \emph{LevelSet R-CNN} which combines the strengths of modern deep learning based Mask R-CNN and classical energy based Chan-Vese level set segmentation framework in an end-to-end manner. In particular, we utilize four heads based on FPN to obtain each detected object, an initial level set, deep robust feature representation for the Chan-Vese energy data terms and a set of instance dependent hyperparameters that balance the energy terms and schedule the optimization procedure. We demonstrated the effectiveness of our method on COCO and Cityscapes showing improvements on both datasets. 